\documentclass[11pt, table]{article}
\pdfoutput=1

\usepackage[preprint]{acl}
\usepackage{booktabs}
\usepackage{times}
\usepackage{latexsym}
\usepackage{multirow}
\usepackage{rotating}

\usepackage[T1]{fontenc}
\usepackage[utf8]{inputenc}
\usepackage{microtype}
\usepackage{inconsolata}

\usepackage{graphicx}
\usepackage{amsmath}
\usepackage{array}
\usepackage{amsfonts}
\usepackage{xcolor}
\usepackage{pgf}
\usepackage{pgfplots}
\usepackage{collcell}
\usepackage{float}
\usepackage{subcaption}
\usepackage{enumitem}
\usepackage{xspace}
\usepackage{lipsum}
\usepackage{hyperref}
\usepackage{cleveref}

\usepackage{duckuments}

\newcommand{\methodname}{\textsc{PartRep}}

\definecolor{winblue}{HTML}{D4E6F1}
\definecolor{lossred}{HTML}{FADBD8}
\newcommand{\win}[1]{\cellcolor{winblue}#1}
\newcommand{\loss}[1]{\cellcolor{lossred}#1}

\definecolor{high}{HTML}{ec462e}
\definecolor{low}{HTML}{ffffff}
\newcommand*{\opacity}{90}
\newcommand*{\MinNumber}{0.0}
\newcommand*{\MidNumber}{7.0}
\newcommand*{\MaxNumber}{15.0}

\newcommand{\colorcell}[1]{
 \pgfmathsetmacro{\value}{#1}

 \pgfmathparse{\value > \MaxNumber ? 1 : 0}
 \ifnum\pgfmathresult=1
  \hspace{-0.33em}\cellcolor{high!\opacity}{#1}
 \else
  \pgfmathparse{\value > \MidNumber ? 1 : 0}
  \ifnum\pgfmathresult=1
   \pgfmathparse{int(round(100*(\value - \MidNumber)/(\MaxNumber - \MidNumber)))}
   \xdef\tempa{\pgfmathresult}
   \hspace{-0.33em}\cellcolor{high!\tempa!yellow!\opacity}{#1}
  \else
   \pgfmathparse{int(round(100*(\MidNumber - \value)/(\MidNumber - \MinNumber)))}
   \xdef\tempa{\pgfmathresult}
   \hspace{-0.33em}\cellcolor{low!\tempa!yellow!\opacity}{#1}
  \fi
 \fi
}

\definecolor{darkblue}{rgb}{0, 0, 0.5}
\hypersetup{colorlinks=true, citecolor=darkblue, linkcolor=darkblue, urlcolor=darkblue}

\title{\methodname{}: Learning What to Repeat for Decoder-only LLMs
}

\newcommand{\aspace}{\hspace{0.7em}}
\newcommand{\bandung}{$^{\heartsuit}$}
\newcommand{\postech}{$^{\spadesuit}$}

\author{
 Andikawati P Widjaja\bandung \aspace
 Yongjun Kim\postech \aspace
 Hyounghun Kim\postech \aspace
 Jaeho Lee\postech \aspace
 \\
 \bandung{}Bandung Institute of Technology \aspace
 \postech{}Pohang University of Science and Technology
}

\begin{document}

\maketitle

\begin{abstract}

While decoder-only LLMs excel at a vast array of natural language tasks, they suffer from an asymmetric information flow induced by causal attention: later tokens are richer in contextual grounding than earlier ones.
A simple and effective remedy is prompt repetition---just appending a second copy of the prompt before generation can redistribute grounding across positions and improve reasoning performance.
However, full repetition of the original prompt doubles the KV cache footprint and quadruples attention cost during prefill, making it impractical for long-context settings.
We propose \textbf{\methodname{}}, a selective augmentation method that appends only the most informative tokens---rather than the entire prompt.
We use token-wise negative log-likelihood (NLL) as a selection signal, motivated by the hypothesis that less predictable tokens are less recoverable from the surrounding context and therefore benefit more from late-position repetition.
To avoid the heavy cost of a full forward pass for scoring, we train a lightweight gate that predicts high-NLL tokens from early-layer hidden states, enabling token selection during mid-prefill via early exit. 
Across eight benchmarks (including MMLU, GSM8K, and RULER) and three model families (Qwen2.5, Llama3.2, Gemma4), \methodname{} retains most of the gains of full repetition while using only 59.4\% of its KV cache and 79.0\% of its prefill FLOPs.

\end{abstract}

\section{Introduction}

Decoder-only large language models (LLMs) have become the dominant architecture in natural language processing, driving significant advancements across a wide range of generative and analytical tasks \citep{huang2023advancing,achiam2023gpt}.
A defining feature of these models is causal attention, which allows each token to attend to all preceding tokens, while masking out the future ones.
Although this design enables effective autoregressive generation, it also induces an asymmetric information flow: later tokens have richer contextual grounding than earlier ones \citep{springer2024repetition, behnamghader2024llm2veclargelanguagemodels}.
This asymmetry has been linked to several documented limitations, including premise order sensitivity in reasoning \citep{chen2024premise}, option ordering bias \citep{pezeshkpour2024large, wei2024unveiling, ok2026lost}, and the ``lost in the middle'' phenomenon \citep{liu2024lost}.
This raises a natural question: How can we redistribute contextual grounding more evenly?

\begin{figure}[t]
 \centering
 \includegraphics[width=1\linewidth,height=4cm]{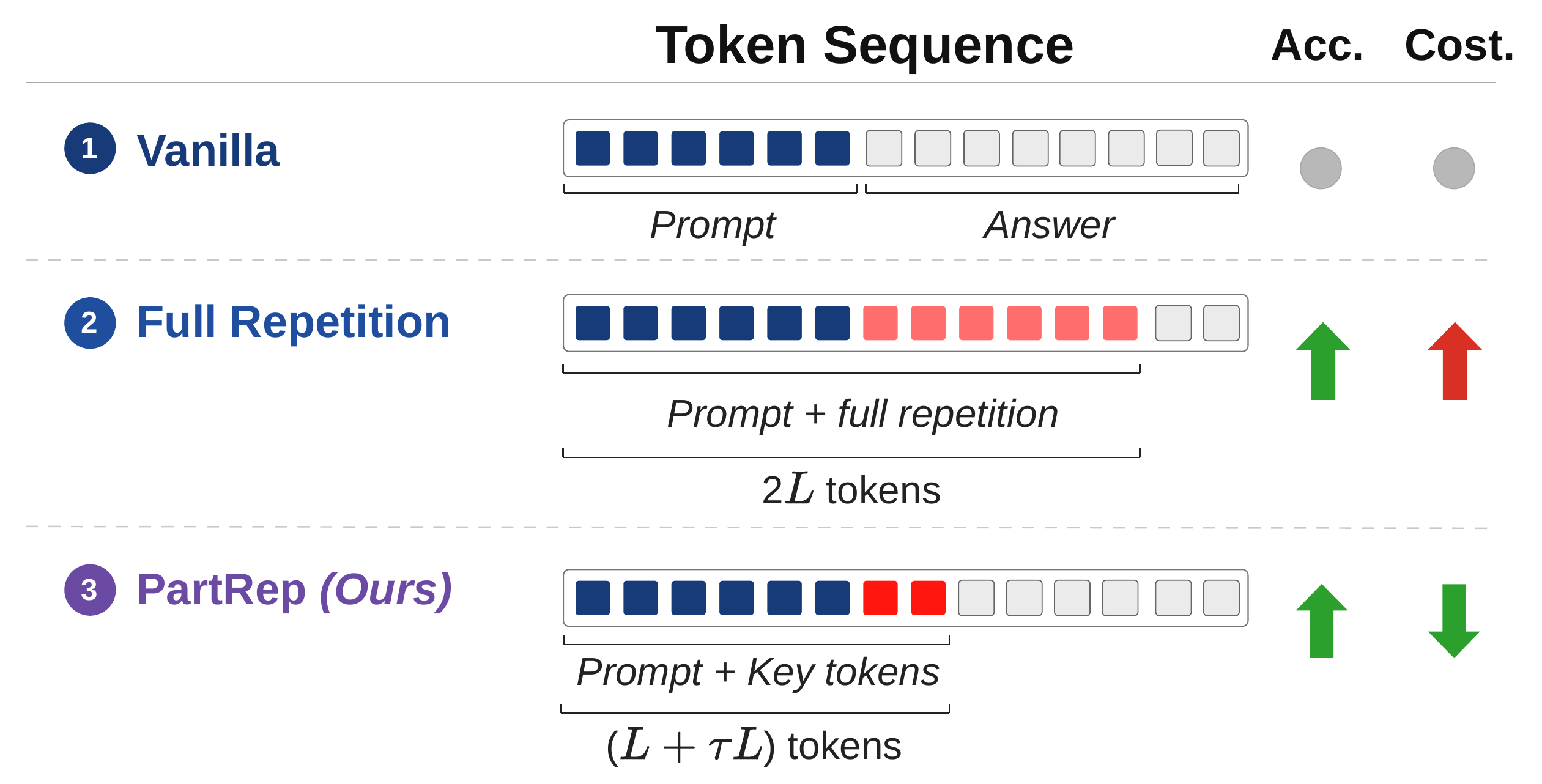}
 \caption{
 \textbf{Overview of token repetition strategies}. 
 Vanilla prompting uses the original prompt once. 
 Full repetition appends the entire prompt, improving accuracy at the cost of doubling the input length to $2L$. 
 \textbf{\methodname{}}, our method, appends only selected key tokens, producing a shorter sequence of length $(L+\tau L)$ that preserves the accuracy gains of repetition while reducing computational cost.
 }
 \label{fig:overview}
\end{figure}

A surprisingly effective remedy for this asymmetry is prompt repetition: present the same prompt twice before generation.
By placing a second copy of the prompt at the end of the input, every original token gains a later ``echo'' that can attend back over the full context, effectively redistributing contextual grounding across the sequence.
Despite requiring no parameter updates or architectural changes, this simple intervention has been shown to improve reasoning performance across diverse tasks \citep{leviathan2025prompt, xu2024re, springer2024repetition}. 

However, it also introduces several limitations: Appending a full copy of the prompt doubles the KV cache footprint and quadruples self-attention FLOPs.
These overheads become increasingly prohibitive as the prompt length grows, rendering this approach unscalable for long contexts. 

To this end, we present \textbf{\methodname{}}, a method to selectively repeat only the most critical tokens---rather than duplicating the entire prompt (Figure~\ref{fig:overview}).
\methodname{} scores each input token by its informational importance, then appends only the highest-scoring tokens before generation.
As the scoring criterion, we adopt the negative log-likelihood (NLL) of each token as a proxy for information density: tokens with high NLL (i.e., high surprisal) are those the model finds least predictable from preceding context, indicating that their content is not redundantly encoded by surrounding tokens.
As we show in~\Cref{subsec:scoringCriterion}, repeating such tokens yields the greatest marginal benefit, as their echoes inject information that would otherwise remain underrepresented in later positions' attention.

Since exact NLL computation requires a full forward pass and would undermine our efficiency goal, we instead train a lightweight gate to predict the top-$\tau$ high-NLL tokens directly from early-layer hidden states, enabling selection during mid-prefill.
The selected tokens are then appended to the original prompt via a short natural-language bridge (e.g., ``\textit{Pay attention to these key tokens...}''), after which a single forward pass produces the final output.

We demonstrate the robustness of our approach across 8 benchmarks, including GSM8K and RULER, and a diverse set of decoder-only LLMs encompassing Qwen2.5, Llama3.2, and Gemma4 families. 
\methodname{} preserves the accuracy benefits of full repetition, while requiring only 59.4\% of the KV cache budget and  79.0\% of the prefill FLOPs.

We summarize our contributions as follows:
\begin{itemize}[leftmargin=*,topsep=0pt,noitemsep]
 \item We propose \methodname{}, a selective prompt augmentation method that approximates the benefits of full repetition at a fraction of its pre-fill compute and KV cache memory cost.
 \item We show that token-wise NLL provides a principled importance signal for token selection and design an efficient gating mechanism that approximates it without a full forward pass.
 \item We empirically validate \methodname{} across eight benchmarks and three model families, showing that it consistently matches or surpasses full repetition across the setups, while substantially reducing prefill computation and memory overhead.
\end{itemize}


\section{Related work}
\label{sec:background}

\paragraph{Prompt repetition.} Several studies have shown that prompt repetition can improve LLM performance. \citet{xu2024re} found that ``re-reading'' a question before chain-of-thought reasoning improves LLM performance, arguing that repetition enables causally masked models to partially emulate bidirectional attention and mitigate limitations of decoder-only architectures. \citet{arora2024just} extended similar ideas to recurrent language models, while \citet{springer2024repetition} formalized the phenomenon through ``echo embeddings,'' showing that representations from repeated tokens substantially outperform single-pass embeddings on retrieval and similarity tasks.
At a larger scale, \citet{leviathan2025prompt} observed consistent gains from prompt repetition across both open- and closed-weight models in non-reasoning mode.
Although prior work has primarily focused on full-prompt repetition, related findings on partial (option-level) repetition by \citet{ok2026lost} further suggest that the benefits that arise from restoring information pathways are blocked by causal attention. 
However, option-level repetition is limited to multiple-choice settings and does not generalize to open-ended generation tasks prevalent in real-world LLM usage.
In contrast, we introduce a task-agnostic framework that selectively repeats only the most informative tokens within an arbitrary prompt, avoiding both the doubled KV cache cost of full repetition and the multiple-choice constraint of option-level repetition.

\paragraph{KV cache eviction.} A number of prior works propose to evict KV cache of \textit{less important} tokens to mitigate the high memory I/O cost of long-context LLM inference. For instance, H$_2$O \citep{zhang2023h2o} retains the tokens that frequently receive high attention during the next-token prediction, while KVzip \citep{kim2025kvzip} asks the model to reconstruct the prompt and utilize the attention during this process. More recent works focus on avoiding computing full attention matrices for each token during the prefill stage to reduce the latency overhead. For example, FastKVzip \citep{kim2026fast} and KVzap \citep{jegou2026kvzap} introduce lightweight predictors trained to predict the token importance based on the hidden states of early transformer layers, where the token importance is estimated by prior methods (e.g., KVzip).

We adopt a similar predictor-based strategy, but for a fundamentally different objective.
Eviction estimates which tokens can be safely discarded, while partial repetition estimates which tokens yield additional benefit when re-attended to, and these two notions of importance are not duals of each other.
We elaborate in \cref{ssec:compare_eviction}, where we show, for example, that the final prompt token is typically essential under eviction yet least useful to repeat.

\paragraph{Summarization.} Another line of work reduces memory cost by compressing the input prompt itself. LLMLingua \citep{jiang2023llmlingua} uses a compact language model to identify and remove \textit{less important} tokens before feeding a summarized prompt to the main model. In this sense, summarization can be viewed as a text-level analogue of KV cache eviction: instead of discarding key-values after prefill, it removes redundant tokens before they enter the model. Our work similarly estimates token importance and constructs a shortened textual representation in the form of a summary. However, the objectives differ fundamentally. LLMLingua assumes that the original prompt contains inherent redundancy and compresses it to reduce memory consumption. In contrast, our goal is not merely to shorten the prompt, but to approximate the accuracy gains of \textit{redundant} full repetition under constrained memory budgets.

\section{Problem formulation}
\label{sec:problem}

Suppose that we are given an initial \textbf{\textit{user prompt}} as input, formulated as a sequence of tokens
\begin{align}
 \mathbf{t} = (t_1, t_2, \dots, t_L)
\end{align}
with length $L$. For decoder-only LLM, the computational complexity of prefill self-attention scales quadratically as $O(L^2)$, and the stored key-value (KV) cache scales linearly as $O(L)$.

The \textbf{\textit{full repetition}} \citep{leviathan2025prompt} is the idea of repeating the full prompt. In other words, the prompt is duplicated to form
\begin{equation}
\mathbf{t}_{\text{FR}} = \mathbf{t} \oplus \mathbf{t},
\label{equation_fullrep}
\end{equation}
where $\oplus$ denotes sequence concatenation. It has been known that prefilling $\mathbf{t}_{\text{FR}}$ enables critical pseudo-bidirectional attention for the sequence, thereby increasing the accuracy of the model.
On the other hand, the prefill complexity gets quadrupled and the KV cache footprint doubles.

Extending this paradigm, we formulate the problem of \textbf{\textit{partial repetition,}} which aims to achieve the benefit of repetition without inheriting its prohibitive memory and latency costs.
Our goal is to isolate an \textit{important, highly informative} subset of the original prompt such that appending only this subset---rather than the full prompt---suffices to improve model accuracy.
Precisely, we consider the prompt structured as
\begin{align}
\mathbf{t}_{\text{PR}} = \mathbf{t} \oplus \mathbf{t}_{\text{part}}
\end{align}
where $\mathbf{t}_{\text{part}}$ denotes the \textit{keyword subsequence} of $\mathbf{t}$. 
Given this prompt structure, our goal is to find a token selector $f(\mathbf{t}) = \mathbf{t}_{\text{part}}$ which maximizes the model accuracy. More concretely, our goal is to solve
\begin{align}
\max_{f}\:\mathrm{acc}(\mathbf{t} \oplus f(\mathbf{t}))
\label{eq:pr_predictor}
\end{align}
where $\mathrm{acc}(Q)$ denotes the accuracy of the base LLM prompted by $Q$, subject to the constraints
\begin{align}
f(\mathbf{t}) \subset \mathbf{t},\qquad |f(\mathbf{t})| \le \tau \cdot |\mathbf{t}|,
\end{align}
where $\subset$ denotes the subsequence relation (instead of a subset), and $\tau \in (0,1)$ is a repetition threshold enforced to meet the given memory budget.

The primary advantage of formulating as a \textit{token selection} (i.e., imposing $f(\mathbf{t}) \subset \mathbf{t}$) is computational. 
The predictor $f$ can be viewed as making a binary decision on each input token, acting as a ``gate'' that accepts or rejects each token. 
As the output dimension is simple, we can expect $f$ to be implementable with a lightweight model.
Indeed, we implement this with a two-layer MLP with attention, on top of the target model's hidden state.
The same may not be possible if we let the output of $f(\mathbf{t})$ lie in the general vocabulary space, e.g., a textual summary of $\mathbf{t}$, or a continuous token space.

\paragraph{Efficiency.} Through compressing the repeated prompt, we can expect the prefill self-attention computation to be proportional to $(1+\tau)^2L^2$, which is lower than $4L^2$ of the full repetition. Likewise, the KV cache will be proportional to $(1+\tau)L$ which is less than $2L$ of the full repetition. 
However, note that partial repetition introduces an additional inference cost for the token selection procedure $f(\cdot)$. Fortunately, as we will see in \Cref{subsec:efficiency}, the cost is small.

\begin{figure*}[t]
 \centering
 \includegraphics[width=\textwidth,height=4cm]{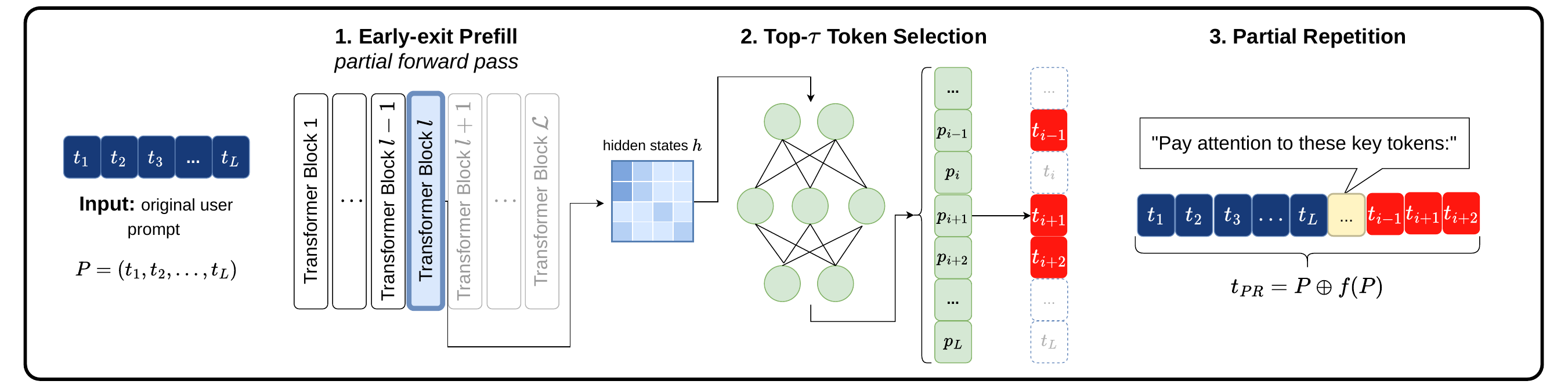}
 \caption{\textbf{Inference procedure of the proposed \methodname{}}. We first prefill the LLM with the original prompt, then pass its early-layer hidden states through the gating module. Next, we select the top-$\tau$ fraction of tokens, and repeat the selected tokens (i.e., append it after the original prompt), then continue with the prefill.}
 \label{fig:method_overview}
\end{figure*}

\subsection{Comparison with KV cache eviction}\label{ssec:compare_eviction}

Recall that the task of KV cache eviction is about deciding ``what to discard'' \citep{zhang2023h2o}. The task of partial repetition, i.e., deciding ``what to repeat,'' is similar in the sense that we need to estimate the importance of each token to reduce the KV cache cost. However, the tasks fundamentally differ from each other in two aspects.

First, tackling partial repetition via KV cache eviction is computationally suboptimal. To see this, consider applying KV cache eviction methods on a fully repeated prompt as a mean of partial repetition. In this case, we must still perform prefill with $2L$ tokens, thus losing any computational advantage in the prefill computation.

Second, the notion of \textit{importance} is different in the two tasks. To see this, consider the last token of the prompt. In partial repetition, this is the least important token to be repeated, as the last token in the original prompt already attended to all other tokens. In KV cache eviction, however, the last token (of the repeated prompt) is typically treated as essential \citep{zhang2023h2o}.

\section{Algorithm}
\label{sec:method}

We develop \textbf{\methodname{}}, a learning-based method that captures the benefits of repetition without inheriting its memory and latency cost by repeating only highly \textit{important} tokens. 

To select these tokens, we first motivate the negative log-likelihood (NLL) as a principled proxy for information density (\Cref{subsec:NLL}).
Since calculating the exact token-level NLL at runtime is prohibitively expensive, we introduce a lightweight gate trained to predict these scores from early-layer hidden states (\Cref{subsec:gating}).
\Cref{subsec:training} describes the offline procedure used to train the gate across diverse domains. At inference, selected tokens are appended to the original prompt before the model's second forward pass.
\Cref{fig:method_overview} illustrates the overall \methodname{} pipeline, with details in \Cref{subsec:inference}.

\subsection{Token importance scoring}
\label{subsec:NLL}

To identify and retain only critical tokens, we use the negative log-likelihood (NLL) of the next-token prediction as a direct proxy of the predicted tokens' information density \citep{jaeger2006speakers}.
In a standard autoregressive language model, the probability of each token $t_i$ conditioned on all preceding tokens $t_{<i}$ is predicted. The NLL for a token is:
\begin{align}
 \text{NLL}(t_i) = -\log P(t_i \mid t_{<i}) \label{eq:nll}
\end{align}
where $P(\cdot|\cdot)$ denotes the next-token probability predicted by the model.
Tokens with high prediction probabilities yield low NLL scores, indicating high redundancy with previous context and low information added.
Conversely, ``surprising'' tokens with low probabilities yield high NLL scores.
Our method leverages this intrinsic property by defining the predictor $f(\cdot)$ to isolate the token subset carrying the highest information density within the repetition threshold $\tau$:
\begin{align}
 f^\star(\mathbf{t}) = \{ t_i \in \mathbf{t} \mid \text{NLL}(t_i) \ge \eta_{\tau} \}. \label{eq:nll_oracle}
\end{align}
Here, $\eta_\tau$ is the threshold determined to meet the required budget for each sample, i.e., $\tau$-quantile.

\paragraph{Rationale.} The NLL-based selector \eqref{eq:nll_oracle} can be viewed as an approximate solution for a more general optimization problem. In particular, consider the following \textit{information maximization} principle:
\begin{align}
\max_{f} H(f(\mathbf{t})|\mathbf{t}),
\end{align}
where $H(\mathbf{t}_2|\mathbf{t}_1)$ denotes the conditional entropy of the model generating a sequence $\mathbf{t}_2$ conditioned on the context $\mathbf{t}_1$.\footnote{Note that this is different from the usual conditional entropy, where we expect $H(f(X)|X) = 0$ for any deterministic $f(\cdot)$. This discrepancy is due to the fact that we consider conditional entropy of a token sequence given its context, modeled by an LLM, rather than modeling $X$ and $f(X)$ as random variables themselves.} In other words, we are selecting a subsequence $f(\mathbf{t})$ that can maximize the information added by the repetition.

\Cref{eq:nll_oracle} approximately solves this problem, where the approximation is that we also condition on the tokens that may be discarded later. Through this approximation, we can save computation; otherwise, we will need to add each context word iteratively, through multiple rounds.

\subsection{Gating mechanism}
\label{subsec:gating}

While $f^\star$ can select critical tokens in a principled way, computing it requires much prefill computation to pass the prompt through the LLM. To alleviate this overhead, we employ a \textit{learning-based approach}: We train a small, lightweight gating module that can approximate $f^\star$ on-the-fly.

Precisely, our gating mechanism works as a composition of two different functions:
\begin{align}
f(\mathbf{t}) = g \circ \phi_{\text{LLM}}(\mathbf{t}). \label{eq:gating_function}
\end{align}
Here, $\phi_{\text{LLM}}(\mathbf{t})$ denotes the hidden states of the original prompt $\mathbf{t}$ extracted from the base LLM at layer $l^\star$, while $g(\cdot)$ is a lightweight token-wise gating function. By leveraging $\phi_{\text{LLM}}(\mathbf{t})$, we can keep $g(\cdot)$ lightweight while minimizing computational overhead. In particular, these features reuse computations from the original prefill stage and therefore incur no additional computational cost.

\paragraph{Architecture.} The token-wise gating function $g(\cdot)$ takes a similar structure with \citet{kim2026fast}, motivated by its success in KV cache eviction. In a nutshell, we first apply a two-layer MLP to map raw hidden states---a highly entangled mixture of syntactic, semantic, and positional signals---onto a compact query space. Then, we use an attention module to compare the query to the keys that represent latent signatures of high-NLL tokens, to compute the final repetition probability.

Concretely, given some hidden state $\mathbf{h} \in \mathbb{R}^d$, we first compute the query
\begin{align}
\mathbf{q} = W_2 \:\mathrm{SiLU}(W_1 \mathbf{h}), \label{eq:twolayer}
\end{align}
where $W_1 \in \mathbb{R}^{2d_{h}\times d}, W_2 \in \mathbb{R}^{d_{h}\times 2d_{h}}$ are weight matrices. We fix $d_{h} = 64$ across models to maintain efficiency. The query $\mathbf{q}$ is then evaluated against $s$ learnable keys $K \in \mathbb{R}^{s \times d_h}$ to output the corresponding weighted sum of values $\mathbf{v} \in \mathbb{R}^s$:
\begin{align}
g(\mathbf{h}) = \sigma\left(\mathbf{v}^\top\mathrm{Softmax}\left(\mathbf{q}K^\top/\sqrt{d_{h}}\right)\right).\label{eq:gating_attention}
\end{align}
Here, $\sigma(\cdot)$ is the sigmoid activation function to generate the final probability score.

\subsection{Training the gating module}
\label{subsec:training}
Now we describe how we train the gating function $f = g \circ \phi_{\text{LLM}}$ (Equation~\ref{eq:gating_function}) using the NLL supervision (Equation~\ref{eq:nll}). To avoid prohibitive training cost, we train only the gating module $g$ and keep the base feature extractor $\phi_{\text{LLM}}$ frozen.

\paragraph{Dataset construction.}
We first construct a training dataset by sampling the prompts $\{\mathbf{t}_i\}_{i=1}^n$ and generating the corresponding feature-NLL pairs
\begin{align}
 \mathcal{D} = \{(\phi_{\text{LLM}}(\mathbf{t}_i),\text{NLL}(\mathbf{t}_i))\}_{i=1}^n,
\end{align}
where $\text{NLL}(\mathbf{t}_i)$ denotes the sequence of token-wise NLL scores. At this stage, we use the NLL itself as the label, instead of $f^\star(\mathbf{t}_i)$. This allows us to reuse the constructed dataset for different repetition thresholds $\eta_\tau$. As this labeling can utilize the batched inference, it can be done efficiently.

For the generality of the learned gating function, we draw sample prompts $\{\mathbf{t}_i\}_{i=1}^n$ from a general educational corpora, rather than domain-specific prompt sets. In particular, we stream and process sequences ranging from 10 to 1024 tokens in length, compiling a robust dataset of 3 million training tokens sampled from FineWeb-Edu (2 million) and Ultra-FineWeb-Edu (1 million).

\begin{table*}[!t]
\centering
\resizebox{\textwidth}{!}{%
\begin{tabular}{l cccccccccc}
\toprule
\textbf{Method} & \textbf{Avg KV Cache} & \textbf{Avg Prefill FLOPs} & ARC & OBQA & MMLU & MedQA & SciQ & MMLU-Pro & GSM8K & \textbf{Avg.} \\
\midrule

No Repetition & 235.5 & 1.549 T & 79.8 & 76.6 & 62.5 & 47.5 & 90.3 & 27.8 & 84.2 & 66.9 \\
\midrule
{\color{gray}\textbf{\textit{Appending summary}}}\\
+ Na\"{i}ve Summary & 277.4 & 3.676 T & \win{81.7} & \loss{76.5} & \loss{60.8} & \win{50.9} & \win{91.6} & \win{28.6} & \win{85.5} & \win{67.9} \\

+ LLMLingua & 277.4  & 2.105 T & \win{81.5} & \win{77.8} & \loss{61.0} & \win{49.3} & \win{90.7} & \loss{27.7} & \loss{84.0} & \win{67.4} \\

\midrule
Full Repetition & 471.1 & 3.140 T & \win{82.8} & \win{78.0} & \loss{60.3} & \win{51.0} & \win{92.9} & \loss{27.3} & \win{87.3} & \win{68.5} \\
\midrule
{\color{gray}\textbf{\textit{Compressing full. rep.}}}\\
+ Echo Eviction & 235.5 & 3.152 T & \win{81.3} & \loss{69.6} & \loss{60.3} & \win{50.7} & \loss{90.1} & \loss{22.0} & \loss{76.6} & \loss{64.4} \\

+ H2O Eviction & 270.8 & 3.162 T & \loss{75.0} & \loss{76.2} & \loss{60.5} & \win{50.6} & \loss{77.7} & \loss{27.4} & \loss{72.9} & \loss{62.9} \\

+ LLMLingua Comp. & 78.6 & 1.130 T & \loss{15.2} & \loss{18.8} & \loss{14.8} & \loss{12.2} & \loss{20.6} & \loss{10.5} & \loss{12.2} & \loss{14.9} \\

\midrule
\methodname{} (ours, $\tau=0.15$) & 280.4 & 2.481 T & \win{81.4} & \win{77.8} & \win{63.2} & \win{50.3} & \win{92.2} & \win{27.8} & \win{85.5} & \win{68.3} \\

\bottomrule
\end{tabular}%
}
\caption{Accuracy comparison of various prompt repetition methods across seven benchmarks, on Qwen 2.5-3B. Colored cells in 
\colorbox{lossred}{red} and \colorbox{winblue}{blue} indicates performance drops and gains relative to the ``No Repetition,'' respectively.}
\label{tab:main_results}
\end{table*}

\paragraph{Training procedure.} Given a sample prompt $\mathbf{t}$ of length $L$, training $g$ is formulated as $L$ token-wise binary classification tasks, where the ground-truth labels are provided by the top-$\tau$ NLL selector $f^\star$ (for a designated $\tau$). Specifically, we generate binary labels as
\begin{align}
\mathbf{y} = \big(\mathbf{1}\{t_1 \in f^\star(\mathbf{t})\},\ldots,\mathbf{1}\{t_L \in f^\star(\mathbf{t})\}\big),
\end{align}
where  $\mathbf{1}$ denotes the indicator function.
Then, we train with the sample-wise loss
\begin{align}
\ell(\mathbf{t}) = \frac{1}{L}\sum_{j=1}^L \lambda_j\: \ell_{\text{CE}}([f(\mathbf{t})]_j,y_j),
\end{align}
where $[\cdot]_j$ denotes the $j$th entry of a vector. Here, $\lambda_j$ is the weight determined to mitigate the effect of class imbalance. In particular, the positive-labeled tokens (i.e., $y_j$ = 1) are upweighted by
\begin{align}
\lambda_j = (1-\tau) / \tau,
\end{align}
representing the ratio between negative and positive tokens in the training set. For the negative-labeled tokens (i.e., $y_j=0$), we simply use $\lambda_j = 1$.

\subsection{Inference}
\label{subsec:inference}
\paragraph{Connecting prompt.} Given an input prompt $\mathbf{t}$, the model performs an initial forward pass and pauses at the target layer $l^\star$. Hidden states at this layer are used to compute $f(\mathbf{t})$, and the selected tokens are then appended to the original prompt, connected through a trigger:
\texttt{``\textbackslash nPay attention to these key tokens:\textbackslash n''}.
The model then performs the full forward pass over the extended prompt $\mathbf{t}\oplus f(\mathbf{t})$, reusing the hidden layer features of the original prompt $\mathbf{t}$. 

\paragraph{Token windowing.} \methodname{} directly appends tokens selected by the gate, which is effective when those tokens alone provide sufficient cues for downstream prediction.
However, token-level repetition can be insufficient in two cases.
First, subword tokenization may split an informative word into multiple fragments, so repeating only one selected fragment yields an incomplete lexical cue.
Second, when the answer depends on local compositional structure, such as relations expressed within a sentence, repeating isolated salient tokens may discard the context needed
to interpret them correctly.

To handle these cases, we introduce token windowing as an optional
post-selection expansion strategy.
When enabled, each selected token can be expanded to include either the full word containing it or a broader local context window.
This option allows \methodname{} to preserve lexical integrity or short-range relational structure when isolated token repetition is insufficient.

\section{Experiment}
\label{sec:experiment}

\subsection{Experimental setup}
\paragraph{Benchmarks.}
We evaluate on a total of eight tasks. Five benchmark datasets assess general knowledge and scientific retrieval capabilities: ARC-Challenge \citep{clark2018think}; OpenBookQA \citep{mihaylov2018can}; MMLU \citep{hendrycks2020measuring}; MedQA \citep{jin2021disease}; SciQ \citep{welbl2017crowdsourcing}. Two benchmark datasets assess more complex reasoning abilities: MMLU-Pro \citep{wang2024mmlu}; GSM8K \citep{cobbe2021training}. One benchmark dataset assesses long-context capabilities: RULER \citep{hsieh2024ruler}.

At the inference stage, we use complete test sets for all benchmarks, except for RULER, where we use a subset of 100 questions per each tasks, 1300 questions in total.

\paragraph{Models.}
We conduct experiments on three different instruction-finetuned LLMs: Qwen 2.5-3B \citep{qwen2024qwen2}; Llama 3.2-3B \citep{grattafiori2024llama}; Gemma 4-E4B \citep{googledeepmind2026gemma4modelcard}.

Notably, Gemma 4-E4B employs a hybrid mechanism alternating between local sliding window and global attention layers.
For this model, we strictly constrain our extraction point to coincide with a global attention layer to prevent an information bottleneck within the extracted hidden states.

\paragraph{Baselines.}
We compare our method against seven standard baselines with repetition. Three of them are methods that do not involve full repetition.
\begin{itemize}[leftmargin=*,topsep=0pt,parsep=0pt,itemsep=0pt]
\item \textit{No repetition.} This is the vanilla setup that uses the user prompt as is without any repetition.
\item \textit{+ Na\"{i}ve summary.} After the original prompt, we append the summary of the prompt generated by the model itself.
\item \textit{+ LLMLingua.} Same, but using LLMLingua instead of summarization \citep{jiang2023llmlingua}.
\end{itemize}
The other four are the methods based on full repetition.
\begin{itemize}[leftmargin=*,topsep=0pt,parsep=0pt,itemsep=0pt]
\item \textit{Full repetition.} We repeat the whole prompt twice \citep{leviathan2025prompt}.
\item \textit{+ Echo eviction.} After processing fully repeated prompt, we remove the hidden states of the first half \citep{springer2024repetition}.
\item \textit{+ H2O eviction.} Same, but we use H2O to select tokens to evict \citep{zhang2023h2o}.
\item \textit{+ LLMLingua.} We apply LLMLingua on the fully repeated prompt \citep{jiang2023llmlingua}.
\end{itemize}

\paragraph{Evaluation metrics.}
Our evaluation focuses on assessing the tradeoffs across three dimensions:
\begin{itemize}[leftmargin=*,topsep=0pt,parsep=0pt,itemsep=0pt]
\item \textit{Accuracy.} Accuracy on the target task.
\item \textit{KV cache.} The average number of prompt tokens whose layerwise KV cache is stored.
\item \textit{Prefill compute.} The computation required for prefilling the prompt, measured in FLOPs.
\end{itemize}
Note that we account for algorithm-specific computational overhead, including prompt processing and, when applicable, the additional cost of token selection, summarization, or cache eviction.

\paragraph{Implementation details.} See Appendix~\ref{app:training_details} for training and Appendix~\ref{app:inference_details} for inference details.

\subsection{Main results}
\label{subsec:results}
Table~\ref{tab:main_results} reports the accuracy of Qwen 2.5-3B under different repetition strategies.
We first observe that full repetition provides the strongest gain with over 1.6\%p increase over vanilla, confirming that repeating the prompt benefits decoder-only LLMs.

\methodname{} closely matches this gain with an average accuracy of 68.3\%, while repeating only a small subset of the prompt. Critically, our method provides consistent gains over benchmarks, even on tasks where full repetition drops the performance, e.g., MMLU and MMLU-Pro. This may be due to the fact our method can help avoid the accuracy degradations from having too long context (e.g., lost-in-the-middle phenomenon).

Summary-based alternatives also perform strongly, but remain below our method. This suggests that compressing prompt retains useful information, but replacing the context with an abstract summary is less effective than preserving the original prompt and appending selected tokens.

\begin{table}[!t]
\centering
\resizebox{\linewidth}{!}{%
\begin{tabular}{lccc}
\toprule
\textbf{Method} & \textbf{Qwen 2.5} & \textbf{Llama 3.2} & \textbf{Gemma 4} \\
\midrule
No Repetition & 62.5 & 54.1 & 74.0 \\
\midrule
+ Na\"{i}ve Summary & \loss{60.8} & \loss{52.7} & \win{74.2} \\
+ LLMLingua & \loss{61.0} & \loss{53.5} & \win{76.1} \\
\midrule
Full Repetition & \loss{60.3} & \win{54.8} & \loss{72.9} \\
\midrule
+ Echo Eviction & \loss{60.3} & \loss{53.2} & \win{74.5} \\
+ H2O Eviction & \loss{60.5} & \loss{52.6} & \loss{72.3} \\
+ LLMLingua Comp.& \loss{14.8} & \loss{11.8} & \loss{24.4} \\
\midrule
\methodname{} (ours) & \win{63.2} & \win{54.3} & \win{75.8} \\

\bottomrule
\end{tabular}%
}
\caption{Accuracy comparison across various LLM architectures, measured on MMLU benchmark.} 
\label{tab:cross_model}
\end{table}

\begin{table}[!t]
\centering
\resizebox{0.8\linewidth}{!}{%
\begin{tabular}{lccc}
\toprule
&
\multicolumn{3}{c}{\textbf{Context Length}} \\
\cmidrule(lr){2-4}
\textbf{Method} & \textbf{4k} & \textbf{8k} & \textbf{16k} \\
\midrule
No Repetition    & 77.7 & 80.0 & 72.3 \\
Full Repetition  & 79.2 & 82.3 & 72.3 \\
LLMLingua     & 69.2 & 73.1 & 63.5 \\
\midrule
\methodname{} (ours, $\tau=0.3$)
        & \textbf{80.0} & \textbf{83.8} & \textbf{73.2} \\
\bottomrule
\end{tabular}%
}
\caption{\textbf{Accuracy comparison on the RULER long-context benchmark at varying context lengths}. For PartRep, we use $\tau=0.3$ with local-context token window expansion.}
\label{tab:ruler}
\end{table}

\subsection{Other LLM architectures}
In \Cref{tab:cross_model}, we provide additional evaluations on diverse LLM architectures with varying baseline capabilities: Llama 3.2-3B and Gemma 4-E4B. In particular, we evaluate on the MMLU benchmark. The results confirm that our method achieves consistent gains over various architectures. Notably, under the strict constraints of Gemma models' hybrid sliding-window and global attention mechanisms, our method stays reliable.

\subsection{Long-context tasks}
We further evaluate our method on long-context settings using the RULER benchmark (\Cref{tab:ruler}) and Qwen2.5-3B. Results confirm that repetition remains beneficial even when the input context becomes substantially longer. 

\methodname{} achieves the strongest performance across all evaluated context length, although the gap gets narrower as the number of token grows. This implies that selectively repeating ``only'' informative tokens remains effective in long-context scenarios. In contrast, full repetition may just repeat redundant information and suffer from the drawbacks of long context, and compressing the prompt (i.e., using LLMLingua) may discard critical structural cues that can only be preserved by keeping the original context intact.

\section{Analysis}\label{sec:analyses}

\subsection{Ablation studies}

\paragraph{Gating module.} In \Cref{tab:architecture}, we compare the effectiveness of the two-layer MLP (Eq.\ref{eq:twolayer}) against alternative architectures: a lighter linear layer and a heavier transformer-based architecture. We use ARC-Challenge benchmark for this comparison. We observe that our architecture achieves a nice trade-off point of accuracy and computation. Simpler models achieve lower accuracy, while the benefits considering more complicated architectures gets saturated over this scale.

\begin{table}[!t]
\centering
\resizebox{0.9\linewidth}{!}{%
\begin{tabular}{lccc}
\toprule
\textbf{Gating module} & \textbf{Accuracy (\%)} & \textbf{Gate FLOPs} \\
\midrule
No Repetition & 79.8 & 0 k \\
\midrule
Linear & 80.7 & 270.6 k \\
2-layer MLP (ours) & \textbf{81.4} & 549.0 k \\
Transformer-based & 81.3 & 654.3 k \\
\bottomrule
\end{tabular}%
}
\caption{\textbf{Ablations on gating module architecture}. Our choice (2-layer MLP) achieves favorable accuracy with minimal computational overhead.}
\label{tab:architecture}
\end{table}

\begin{table}[t]
\centering
\resizebox{\linewidth}{!}{%
\begin{tabular}{lcc}
\toprule
\textbf{Window Expansion} & \textbf{Accuracy} & \textbf{KV cache} \\
\midrule

No windowing & 61.5 & 90.1 \\
1 word & 64.6 & 94.5 \\
+1 word (left), +1 word (right) & 66.0 & 110.2 \\

\bottomrule
\end{tabular}
}
\caption{
\textbf{Effect of token-window expansion for handling fragmented subword
selections}. Adding local word context improves accuracy from 61.5 to
66.0 while introducing only a moderate KV-cache increase.
}
\label{tab:token_windowing}
\end{table}

\paragraph{Token windowing.} We also ablate the token windowing on the Llama 3.2-3B, which uses subword tokenization. 
\Cref{tab:token_windowing} reports the effect of expanding selected token span on ARC-Challenge.
As shown in Table~\ref{tab:token_windowing},
whole-word expansion improves accuracy from 61.5 to 64.6 with a small
increase in KV cache, while adding one neighboring word on each side further improves accuracy to 66.0 with a moderate increase in KV cache.

\paragraph{Other ablations.} We provide more ablations on:
\begin{itemize}[leftmargin=*,topsep=0pt,parsep=0pt,itemsep=0pt]
\item The scoring criterion for $f^\star$ (Appendix~\ref{subsec:scoringCriterion})
\item The early exit layer index $l^\star$ (Appendix~\ref{subsec:layerChoice})
\item The repetition budget $\tau$ (Appendix~\ref{RepetitionBudget})
\item The connecting prompt (Appendix~\ref{subsec:PromptTrigger})
\item The number of learnable keys $s$ (Appendix~\ref{subsec:learnableKeys})
\end{itemize}

\begin{figure}[t]
 \centering
 \includegraphics[width=0.9\linewidth]{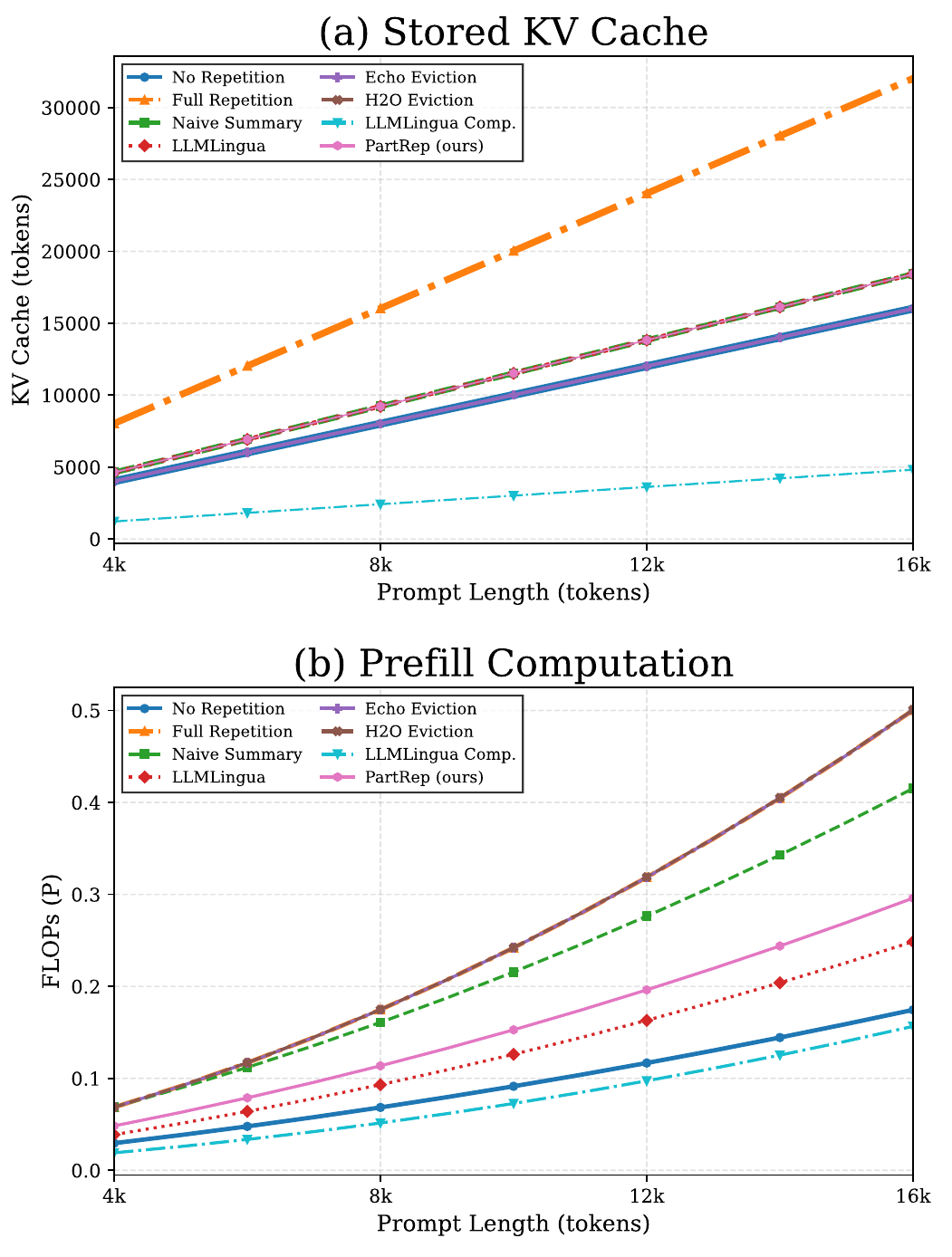}
 \caption{\textbf{Memory and compute scaling with prompt length}. 
 The panel (a) reports the number of stored KV cache tokens, and the panel (b) reports the estimated prefill FLOPs.
 As prompt length increases, Full Repetition incurs substantially larger overhead, whereas Partial Repetition maintains lower KV cache usage and prefill compute by appending only selected informative tokens.}
 \label{fig:tradeoff}
\end{figure}

\subsection{Efficiency comparison}
\label{subsec:efficiency}

\Cref{fig:tradeoff} exhibits the KV cache and FLOPs required by each baseline as prompt length scales. While full repetition maximizes reasoning accuracy, the curve shows it scales prohibitively, doubling the KV cache footprint and inflating prefill complexity to $O(4L^2)$ FLOPs. Meanwhile, as the prompt gets longer, \methodname{}'s KV cache growth turns more and more negligible. As for the complexity, although the gate's token selection procedure introduces additional computational overhead, our empirical results demonstrate that the total required FLOPs still remain substantially lower than what is required in full repetition.

\section{Conclusion}
We propose \methodname{}, a selective prompt augmentation method that approximates the benefits of full repetition at a fraction of its compute and memory cost.
Across eight benchmarks and three model families, \methodname{} consistently improves over vanilla prompting and remains competitive with full repetition across reasoning, knowledge, and long-context tasks, requiring only 59.4\% of the KV cache footprint and 79\% of the prefill FLOPs of the full repetition.
Our ablation studies further demonstrate that high-NLL token supervision, lightweight gating, and local token-window expansion are effective design choices to balance accuracy and inference efficiency.
We believe that \methodname{} provides a practical approach for improving decoder-only LLM inference, particularly in long-context scenarios where efficient use of KV cache and prefill computation is essential.

\section*{Limitations}
One limitation of the \methodname{} is that it requires an offline training for each target LLM, instead of being applicable zero-shot or transferrable across architectures. This undermines the applicability of our framework to scenarios where we do not have much training budget.

Another key limitation is its applicability to different data modalities, e.g., images. Our approach relies on an implicit assumption that each token can convey a meaningful signal by itself. However, as repeating only a few tokens from an image and concatenating them as a sequence may produce a highly distorted image, our method may not generalize well to vision-language models.

\bibliography{main}

\clearpage

\appendix
\section{Implementation Details}\label{app:ImplementationDetails}

\subsection{Training Details}\label{app:training_details}
We train our gate for 25 epochs using Adam optimizer with learning rate of $1\mathrm{e}{-3}$ and batch size of 4096.
We also apply a CosineAnnealingLR scheduler and use early stopping based on validation loss with a patience of 5 epochs.

\subsection{Inference Details}\label{app:inference_details}
All experiments are performed with greedy decoding single runs and bfloat16 precision. Maximum output tokens are set to 1000 for most of the benchmarks, except GSM8K, which allows up to 3000. We infer using vLLM on 6 out of 8 prompt repetition methods, except for Echo Eviction and H2O Eviction which use manual KV decoding. 

\section{Additional Ablations}
This appendix provides additional analyses of the design choices behind \methodname{}.
We study five components of the method: the target scoring
criterion for token selection, the early-exit layer used for token scoring, the repetition budget, the connector prompt used to append selected tokens, and the effect of the number of learnable keys.
Unless otherwise specified, each ablation varies only one component
while keeping all remaining settings fixed to the default configuration used in the main experiments.
We conduct all ablation experiments on ARC-Challenge, with Qwen2.5-3B.

\subsection{Token Scoring Criterion}
\label{subsec:scoringCriterion}
\begin{figure}[H]
 \centering
 \includegraphics[width=\linewidth]{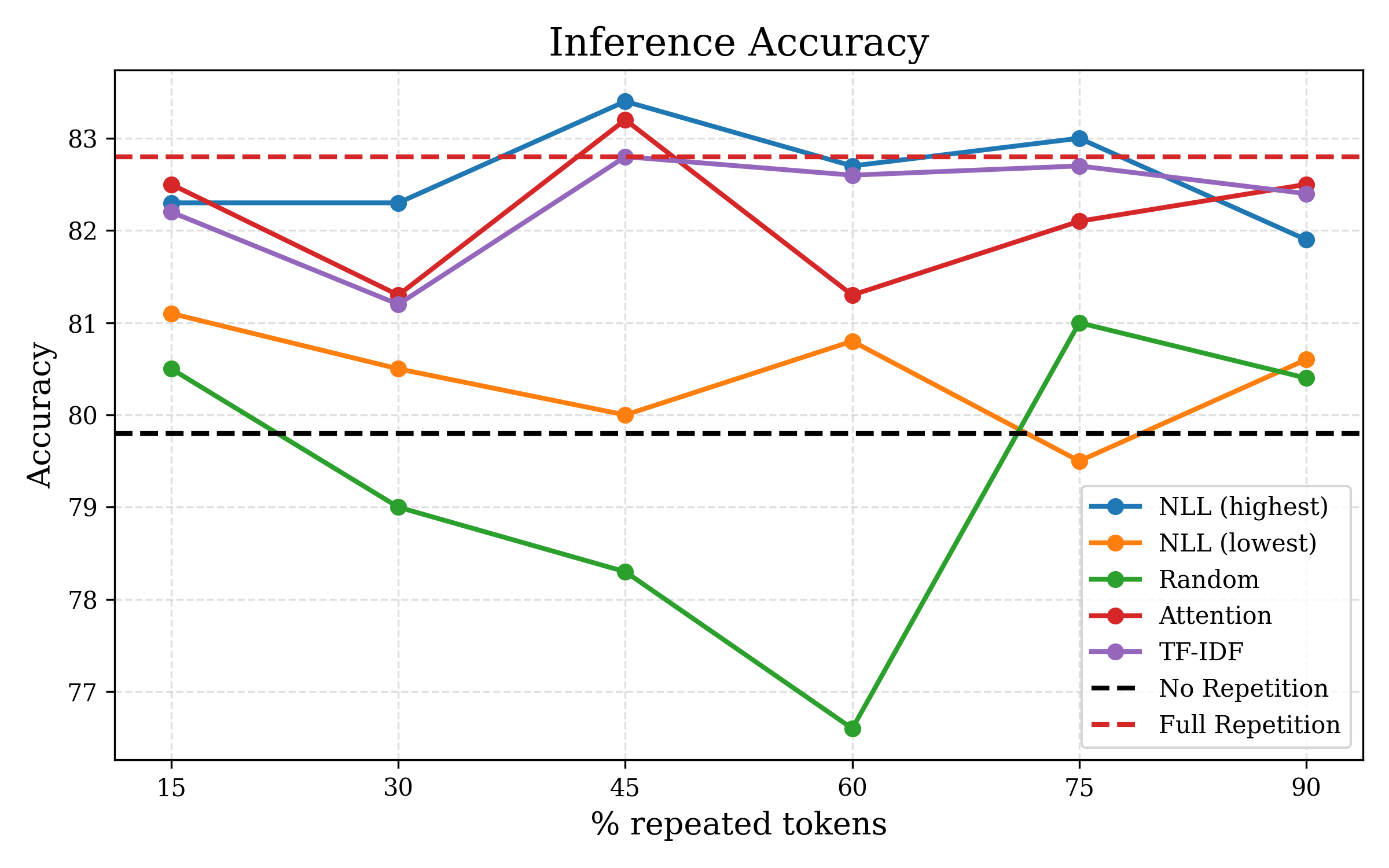}
 \caption{\textbf{Comparison of token scoring criteria under different repetition scoring criteria}.
 Selecting tokens with the highest NLL consistently yields strong
performance, supporting NLL as an effective supervision signal for the gating module.}
 \label{fig:criterion}
\end{figure}
As our method uses token-wise negative log-likelihood (NLL) as the target importance signal for training the gating module, we first study whether this is a valid choice compared to several considerable choices. 
We compare the highest NLL option with low-NLL tokens, random selection, attention-based selection, and TF--IDF-based selection.
For TF--IDF-based selection, we use training dataset to measure the IDF, and target prompt to measure the TF.

Figure~\ref{fig:criterion} shows that selecting high-NLL tokens provides the most reliable accuracy improvements across repetition budgets and achieves the strongest overall performance. In contrast, repeating low-NLL tokens or randomly selected tokens is substantially less effective, indicating that the benefit does not arise merely from adding extra tokens.
Attention and TF-IDF scores show promising results, but we choose high-NLL since they are a more principled solution as we have shown in~\Cref{subsec:NLL}, and they consistently match or surpass those scores.
These results support our hypothesis: tokens that are difficult to predict from preceding context contain information that is particularly valuable to reintroduce at later positions.

\begin{table}[t]
 \centering
 \footnotesize
 \setlength{\tabcolsep}{0pt}
 \renewcommand{\arraystretch}{1.08}

 \begin{tabular}{@{}p{0.36\columnwidth}@{\hspace{0.02\columnwidth}}p{0.60\columnwidth}@{}}
  \toprule
  \textbf{Connector Prompt} & \textbf{Example Query} \\
  \midrule

  Verbal &
  \texttt{"\textbackslash nPay attention to these}
  \newline
  \texttt{key tokens:\textbackslash n\{tokens\}"} \\
  \addlinespace[2pt]

  ChatML &
  \texttt{"\textbackslash n$<$|im\_start|$>$\{tokens\}}
  \newline
  \texttt{$<$|im\_end|$>$"} \\
  \addlinespace[2pt]

  Verbal+ChatML &
  \texttt{"\textbackslash n$<$|im\_start|$>$Pay}
  \newline
  \texttt{attention to these key tokens:}
  \newline
  \texttt{\textbackslash n\{tokens\}$<$|im\_end|$>$"} \\
  \addlinespace[2pt]

  Raw &
  \texttt{"\textbackslash n\{tokens\}"} \\
  \addlinespace[2pt]

  XML &
  \texttt{"\textbackslash n$<$keyword$>$\{tokens\}}
  \newline
  \texttt{$<$/keyword$>$"} \\
  \addlinespace[2pt]

  Verbal (Technical) &
  \texttt{"\textbackslash nTokens with high}
  \newline
  \texttt{Negative Log-Likelihood:}
  \newline
  \texttt{\textbackslash n\{tokens\}"} \\

  \bottomrule
 \end{tabular}

 \caption{
  Connector prompt templates are evaluated for appending the selected
  token subsequence to the original prompt.
 }
 \label{tab:ablation_verbose}
\end{table}

\subsection{Early-exit Layer Choice}
\label{subsec:layerChoice}
\begin{figure}[h]
  \centering
  \includegraphics[width=\linewidth]{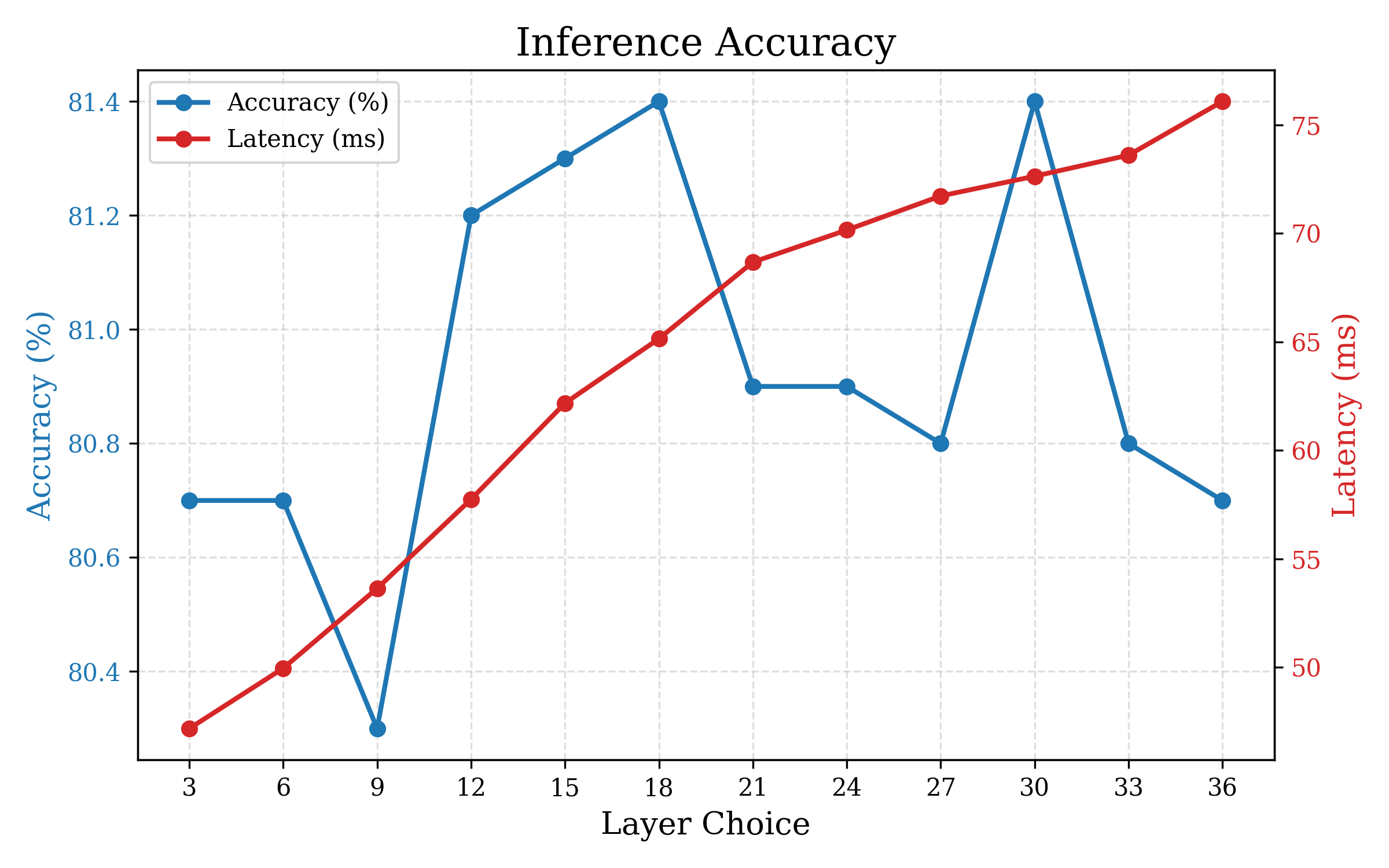}
  \caption{\textbf{Effect of the early-exit layer used by the gating module}. 
  Accuracy improves when the gate receives sufficiently contextualized hidden states, whereas latency increases monotonically with extraction depth.
  We select layer 18 as the default operating point.
 }
\label{fig:ablation_layer}
\end{figure}
We study which intermediate layer should be used to extract the
hidden states consumed by the gating module. 
Extracting features from very shallow layers reduces selection latency, but these representations may not yet contain sufficient semantic information for identifying high-information tokens.
Conversely, extracting from deeper layers increases latency and reduces the computational advantage of early exit.

Figure~\ref{fig:ablation_layer} shows that accuracy improves substantially once moderately deep representations are used, while later extraction points provide limited or no additional benefit despite higher latency.
In particular, layer 18 achieves the best accuracy--latency trade-off.
We therefore use this layer as the default extraction point in the main experiments.

\subsection{Repetition Budget}
\label{RepetitionBudget}
\begin{figure}[H]
 \centering
 \includegraphics[width=\linewidth]{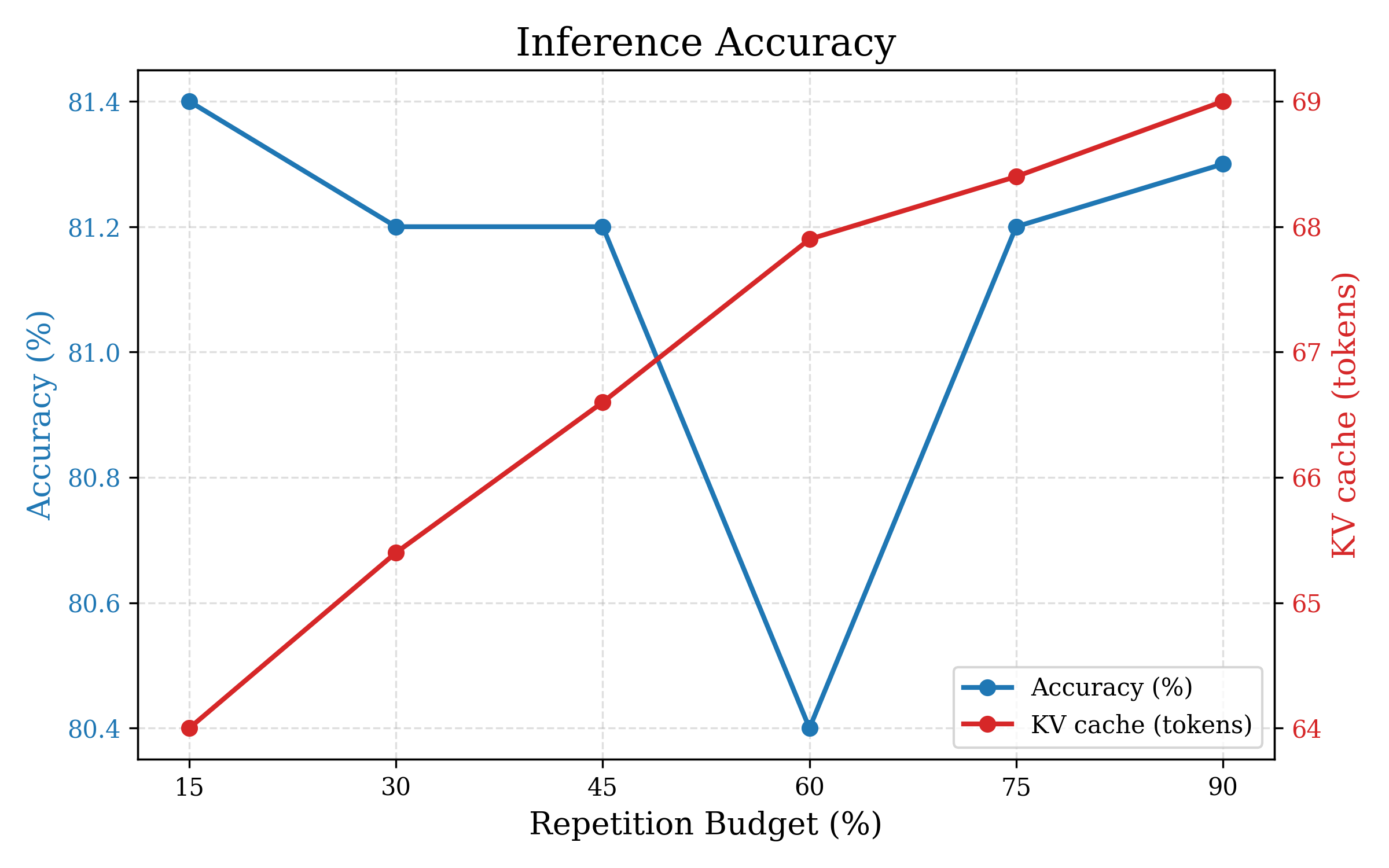}
 \caption{\textbf{Effect of the repetition ratio $\tau$}. Larger budgets increase the KV-cache footprint, while accuracy is non-monotonic, suggesting that selectively repeating a compact set of informative tokens is preferable to indiscriminately increasing the repeated context.}
 \label{fig:repetition_budget}
\end{figure}
We next examine the effect of the repetition budget $\tau$, which
controls the fraction of original prompt tokens appended by Partial
Repetition.
Increasing $\tau$ provides the model with more repeated information, but also increases the KV-cache footprint and prefill cost.
Thus, the optimal operating point should preserve the benefit of
repetition without approaching the overhead of full repetition.

As shown in Figure~\ref{fig:repetition_budget}, accuracy does not improve monotonically with a larger repetition budget.
We conjecture that this non-monotonic trend is partly associated with the verbal prompt trigger used to introduce the repeated tokens.
The trigger frames the appended subsequence as a compact set of salient cues; however, as $\tau$ increases, the repeated subsequence may contain increasingly redundant or weakly informative tokens, making this framing less precise and diluting the salience of genuinely useful tokens.

The selected operating point, $\tau=0.15$, achieves strong accuracy while retaining a substantially smaller KV-cache footprint than full repetition.

\subsection{Connecting Prompt}
\label{subsec:PromptTrigger}
\begin{figure}[H]
 \centering
 \includegraphics[width=\linewidth]{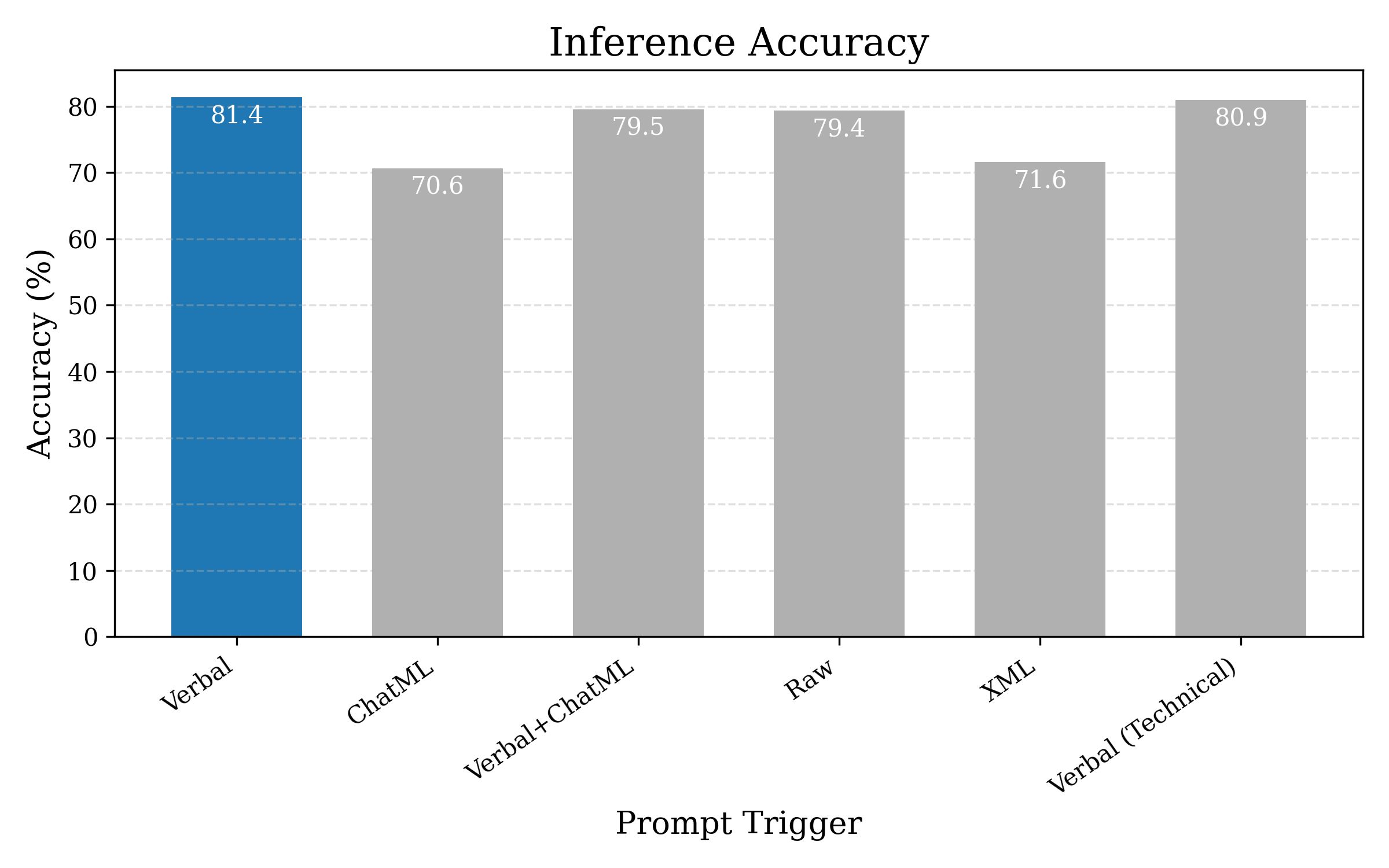}
 \caption{\textbf{Comparison of connector prompts used to append the selected tokens in ARC}. A simple natural-language instruction, ``Pay attention to these key tokens:'', performs best and is used as the default connector.}
 \label{fig:ablation_verbose}
\end{figure}
After selecting informative tokens, Partial Repetition appends them to the original prompt through a short connector string. 
We evaluate whether the form of this connector affects the usefulness of the repeated tokens. 
Specifically, we compare a natural-language verbal instruction, structured wrappers such as ChatML and XML, a raw token
concatenation baseline, and a more technical verbal instruction. See example query in table~\ref{tab:ablation_verbose}.

Figure~\ref{fig:ablation_verbose} shows that the simple verbal connector, ``Pay attention to these key tokens:'', achieves the strongest accuracy.
This suggests that explicitly framing the appended subsequence as
task-relevant information helps the model interpret the repeated tokens, whereas raw concatenation or unnecessarily structured wrappers are less effective.
We therefore adopt the verbal connector throughout the main experiments.

\subsection{Number of learnable keys.}

\begin{table}[t]
\centering
\resizebox{\linewidth}{!}{%
\begin{tabular}{lcc}
\toprule
\textbf{Num. learnable keys} & \textbf{Accuracy (\%)} & \textbf{Gate FLOPs} \\
\midrule
No Repetition & 79.8 & 0 k \\
\midrule
32 & 80.1 & 544.8 k \\
64 (\textbf{ours}) & \textbf{81.4} & 549.0 k \\
128 & 80.6 & 557.3 k \\
\bottomrule
\end{tabular}%
}
\caption{\textbf{Ablations on the number of learnable keys used for the gating module}. Our choice (64) achieve better accuracy than using less (32) or more (128) keys.}
\label{tab:ablation_keys}
\end{table}

\label{subsec:learnableKeys}
In \Cref{tab:ablation_keys}, we test various choices on the number of learnable keys for the gating module (Eq.~\ref{eq:gating_attention}), on the ARC-Challenge dataset. We observe that our choice of 64 achieves the sweet spot, where increasing or decreasing it only degrades the accuracy.

\section{Qualitative results}

This appendix provides qualitative examples of how various prompt repetition methods reconstruct the final input prompt given to the model. We use examples from the ARC-Challenge dataset and run them on Qwen2.5-3B.
For KV cache eviction method, we represent tokens with evicted {\color{red}{red}}. The original prompt is as follows.

\noindent
\begin{tabular}{|p{\dimexpr\columnwidth-2\tabcolsep-2\arrayrulewidth\relax}|}
\hline
Farmers in Wyoming were concerned because some of their chickens were being preyed upon by hawks that lived in areas around their ranches. The farmers grouped together and hunted the hawks until they were no longer in their area. Which would most likely happen next? \\
A. The chicken population would go down. \\
B. Populations of mice and rats would increase. \\
C. Another bird of prey would replace the hawk. \\
D. The chickens would have a lower rate of disease.\\
Reply with one letter in the format:\\
The answer is: \\
\hline
\end{tabular}

\subsection{No Repetition}
\label{subsec:no_repetition}
This method utilizes the vanilla setup, where the original ARC prompt is given to the model without any additional compressed context or repetition.

\subsection{Full Repetition}
\label{subsec:full_repetition}
This method repeats the entire original prompt twice. It gives the model a complete second look over the prompt, but at the cost of doubling the prompt length.

\noindent
\begin{tabular}{|p{\dimexpr\columnwidth-2\tabcolsep-2\arrayrulewidth\relax}|}
\hline
Farmers in Wyoming were concerned because some of their chickens were being preyed upon by hawks that lived in areas around their ranches. The farmers grouped together and hunted the hawks until they were no longer in their area. Which would most likely happen next? \\
A. The chicken population would go down. \\
B. Populations of mice and rats would increase. \\
C. Another bird of prey would replace the hawk. \\
D. The chickens would have a lower rate of disease.\\
Reply with one letter in the format:\\
The answer is: \\
\hline
\end{tabular}

\subsection{PartRep (Ours)}
This method keeps the original prompt in Section~\ref{subsec:no_repetition} intact and appends only \textit{important} tokens chosen by our method at the end.

\noindent
\begin{tabular}{|p{\dimexpr\columnwidth-2\tabcolsep-2\arrayrulewidth\relax}|}
\hline
Pay attention to these key tokens:\\
Wyoming were concerned some chickens prey grouped hunted until would Pop mice Another replace\\
Reply with one letter in the format:\\
The answer is:\\
\hline
\end{tabular}

\subsection{Na\"{i}ve Summary}
This method appends a short summary generated by the model to the original prompt in Section~\ref{subsec:no_repetition}. Summary length is restricted to $\tau$ tokens, keeping it on a par with \methodname{} in terms of KV-cache.

\noindent
\begin{tabular}{|p{\dimexpr\columnwidth-2\tabcolsep-2\arrayrulewidth\relax}|}
\hline
Here is the summary of the prompt:\\
Farmers in Wyoming are concerned about hawks preying on their chickens, \\
\hline
\end{tabular}

\subsection{LLMLingua}
This method also appends a short summary to the original prompt in Section~\ref{subsec:no_repetition}, however, the prompt is compressed by Microsoft's LLMLingua instead of the model itself. Summary length is also restricted to $\tau$ to keep fair comparison.

\noindent
\begin{tabular}{|p{\dimexpr\columnwidth-2\tabcolsep-2\arrayrulewidth\relax}|}
\hline
Here is the summary of the prompt:\\
Wyoming chickens preyed..?\\
chicken.\\
mice.\\
\hline
\end{tabular}

\subsection{LLMLingua Comp.}
This method compresses the fully repeated prompt in Section~\ref{subsec:full_repetition} using LLMLingua. We only feed the compressed prompt to the main model, hence KV-cache memory remains low.

\noindent
\begin{tabular}{|p{\dimexpr\columnwidth-2\tabcolsep-2\arrayrulewidth\relax}|}
\hline
Wyoming chickens preyed..?\\
chicken.\\
mice rats.\\
..\\
..\\
preyed..?\\
chicken population.\\
mice rats.\\
..\\
..\\
Reply with one letter in the format:\\
The answer is:\\
\hline
\end{tabular}

\subsection{Echo Eviction}
This method first applies Full Repetition as in Section~\ref{subsec:full_repetition}, then evicts the KV-cache that corresponds with the original prompt (Section~\ref{subsec:no_repetition}). Therefore, the textual prompt given to the model is identical to Full Repetition, yet only the repeated part is retained for answer generation.

\noindent
\begin{tabular}{|p{\dimexpr\columnwidth-2\tabcolsep-2\arrayrulewidth\relax}|}
\hline
\color{red}{Farmers in Wyoming were concerned because some of their chickens were being preyed upon by hawks that lived in areas around their ranches. The farmers grouped together and hunted the hawks until they were no longer in their area. Which would most likely happen next?} \\
\color{red}{A. The chicken population would go down.} \\
\color{red}{B. Populations of mice and rats would increase.} \\
\color{red}{C. Another bird of prey would replace the hawk.} \\
\color{red}{D. The chickens would have a lower rate of disease.}\\
\color{red}{Reply with one letter in the format:}\\
\color{red}{The answer is:} \\
Farmers in Wyoming were concerned because some of their chickens were being preyed upon by hawks that lived in areas around their ranches. The farmers grouped together and hunted the hawks until they were no longer in their area. Which would most likely happen next? \\
A. The chicken population would go down. \\
B. Populations of mice and rats would increase. \\
C. Another bird of prey would replace the hawk. \\
D. The chickens would have a lower rate of disease.\\
Reply with one letter in the format:\\
The answer is: \\
\hline
\end{tabular}

\subsection{H2O Eviction}
This method also starts from the fully repeated prompt as in Section~\ref{subsec:full_repetition}, but further applies H2O to retain only selected heavy-hitter tokens in the KV-cache. Number of retained tokens is set to ($L+\tau$) to match the length of \methodname{}.

\noindent
\begin{tabular}{|p{\dimexpr\columnwidth-2\tabcolsep-2\arrayrulewidth\relax}|}
\hline
Farmers in Wyoming were concerned because some of their chickens were being preyed upon by hawks that lived in areas around their ranches. The farmers grouped together and hunted the hawks until {\color{red}{they were no}} longer in {\color{red}{their area}}. Which would most likely happen next? \\
A. The chicken population would \color{red}{go} down. \\
B. Populations of mice and rats would increase. \\
C. Another bird \color{red}{of} prey \color{red}{would} replace \color{red}{the} hawk. \\
D. The chickens would have a lower \color{red}{rate of} disease.\\
Reply with one letter in the format:\\
The answer is: \\
Farmers in Wyoming were concerned because some of their chickens {\color{red}{were being preyed upon by}} hawks . The farmers grouped together and hunted the hawks until {\color{red}{they were no longer in their area}}. Which {\color{red}{would most likely happen next?}} \\
A. {\color{red}{The chicken population would go down.}} \\
B. {\color{red}{Populations of mice and rats would increase.}} \\
C. {\color{red}{Another bird of prey would replace the hawk.}} \\
D. {\color{red}{The chickens would have a lower rate of disease.}}\\
Reply with {\color{red}{one letter in the format}}:\\
The answer is: \\
\hline

\end{tabular}

\section{Resources}
All experiments are performed on 8×NVIDIA GeForce RTX 4090 GPUs.

\section{Artifact Licenses and Intended Use}

In this section, we provide the licenses of the used open-weight models, datasets, and benchmarks that are used in our study.
Our use of these artifacts follows their original licenses and is entirely consistent with their intended use for academic research.

\subsection{Models}
\begin{itemize}[leftmargin=*,topsep=0pt,parsep=0pt,itemsep=0pt]
 \item Qwen2.5-3B Instruct: Qwen RESEARCH LICENSE AGREEMENT
 \item Llama3.2-3B Instruct: Llama 3.2 Community License Agreement
 \item Gemma4-E4B it: Apache License 2.0
 \item llmlingua-2-xlm-roberta-large-meetingbank: MIT License
\end{itemize}

\subsection{Datasets and Benchmarks}
\begin{itemize}[leftmargin=*,topsep=0pt,parsep=0pt,itemsep=0pt]
 \item ARC: Creative Commons Attribution Share Alike 4.0
 \item OpenbookQA: Apache License 2.0
 \item SciQ: Creative Common Attribution Non Commercial 3.0
 \item MedQA: Creative Commons Attribution 4.0
 \item MMLU: MIT License
 \item MMLU-Pro: MIT License
 \item GSM8K: MIT License
 \item Nvidia RULER: Apache License 2.0
 \item FineWeb-Edu: Open Data Commons License Attribution family
 \item Ultra-FineWeb-EDU: Apache License 2.0
\end{itemize}

\end{document}